\documentclass[runningheads]{llncs}
\usepackage[T1]{fontenc}
\usepackage{graphicx}
\usepackage{amsmath,amsfonts}
\usepackage{algorithmic}
\usepackage{textcomp}
\usepackage{xcolor}
\usepackage{xspace}
\usepackage{hyperref}
\usepackage{listings}
\usepackage{enumitem}
\usepackage{multirow}
\usepackage{booktabs}
\usepackage{array}
\usepackage[labelfont=bf]{caption}
\usepackage{subcaption}
\usepackage{pifont}
\usepackage{fancyvrb}
\usepackage{optidef}

\usepackage[ruled,linesnumbered]{algorithm2e}

\usepackage[acronym]{glossaries}

\usepackage{orcidlink}
\usepackage{tcolorbox}

\hypersetup{%
  colorlinks=true,
  linkcolor=black,
  citecolor=black,
  urlcolor=blue
}

\newacronym{OWASP}{OWASP}{Open Web Application Security Project}
\newacronym{CRS}{CRS}{Core Rule Set}
\newacronym{regex}{regex}{regular expression}
\newacronym{WAF}{WAF}{Web Application Firewall}
\newacronym{ML}{ML}{machine learning}
\newacronym{RL}{RL}{Reinforcement Learning}
\newacronym{AI}{AI}{Artificial Intelligence}
\newacronym{NLP}{NLP}{Natural Language Processing}
\newacronym{SVM}{SVM}{Support Vector Machine}
\newacronym{CNN}{CNN}{Convolutional Neural Networks}
\newacronym{LR}{LR}{Logistic Regression}
\newacronym{RF}{RF}{Random Forest}
\newacronym{SQLi}{SQLi}{SQL injection}
\newacronym{XSS}{XSS}{cross-site scripting}
\newacronym{RCE}{RCE}{remote code execution}
\newacronym{PL}{PL}{Paranoia Level}
\newacronym{TNR}{TNR}{True Negative Rate}
\newacronym{FPR}{FPR}{False Positive Rate}
\newacronym{TPR}{TPR}{True Positive Rate}
\newacronym{ROC}{ROC}{Receiver Operating Characteristic}
\newacronym{DR}{DR}{Detection Rate}
\newacronym{API}{API}{Application Programming Interface}
\newacronym{PGD}{PGD}{projected gradient descent}
\newacronym{FGSM}{FGSM}{fast gradient sign method}
\newacronym{XAI}{XAI}{eXplainable Artificial Intelligence}

\graphicspath{{figures/}}

\newcolumntype{C}[1]{>{\centering\let\newline\\\arraybackslash\hspace{0pt}}m{#1}}

\makeatletter
\AtBeginDocument{\let\c@listing\c@lstlisting}
\makeatother

\newcommand{\vct}[1]{\ensuremath{\boldsymbol{#1}}}

\newcommand{\set}[1]{\ensuremath{\mathcal{#1}}}
\newcommand{\con}[1]{\ensuremath{\mathsf{#1}}}

\newcommand{\mycomment}[1]{}

\newcommand{\myparagraph}[1]{\noindent \textbf{#1}}

\newcommand{\mysubparagraph}[1]{\noindent \underline{\textit{#1}}}
\newcommand{\ie}{i.e.\xspace}

\newcommand{\etal}{\emph{et al.}\xspace}

\newcommand{\mlms}{ModSec-Learn\xspace}
\newcommand{\modsecurity}{ModSecurity\xspace}
\newcommand{\crs}{CRS\xspace}

\begin{document}
\title{ModSec-Learn: Boosting ModSecurity with Machine Learning}
\titlerunning{ModSec-Learn: Boosting ModSecurity with Machine Learning}
%
\author{
Christian Scano\inst{1, 2}\orcidlink{0009-0003-7756-6125} \and
Giuseppe Floris\inst{1}\orcidlink{0009-0007-7000-5260} \and
Biagio Montaruli\inst{3,6}\orcidlink{0009-0002-6870-8075} \and \newline
Luca Demetrio\inst{4}\orcidlink{0000-0001-5104-1476} \and
Andrea Valenza\inst{5}\orcidlink{0000-0001-7771-2485} \and
Luca Compagna\inst{3}\orcidlink{0009-0003-1072-4352} \and
Davide Ariu\inst{2}\orcidlink{0000-0001-7970-5959} \and \newline
Luca Piras\inst{2}\orcidlink{0000-0002-2130-9598} \and
Davide Balzarotti\inst{6}\orcidlink{0000-0001-5957-6213} \and
Battista Biggio\inst{1,2}\orcidlink{0000-0001-7752-509X}
}

\authorrunning{C. Scano et al.}
%
\institute{
University of Cagliari, Cagliari, Italy \\
\email{\{battista.biggio,giuseppe.floris\}@unica.it} \and
Pluribus One, Cagliari, Italy \\
\email{\{davide.ariu,luca.piras\}@pluribus-one.it} \and
SAP Security Research, Mougins, France \\
\email{\{biagio.montaruli,luca.compagna\}@sap.com} \and
University of Genova, Genova, Italy \\
\email{luca.demetrio@unige.it} \and
Prima Assicurazioni, Milano, Italy \\
\email{andrea.valenza@prima.it} \and
EURECOM, Biot, France \\
\email{davide.balzarotti@eurecom.fr}
}
\maketitle              
\begin{abstract}
ModSecurity is widely recognized as the standard open-source Web Application Firewall (WAF), maintained by the OWASP Foundation. It detects malicious requests by matching them against the Core Rule Set (CRS), identifying well-known attack patterns. 
Each rule is manually assigned a weight based on the severity of the corresponding attack, and a request is blocked if the sum of the weights of matched rules exceeds a given threshold.
However, we argue that this strategy is largely ineffective against web attacks, as detection is only based on heuristics and not customized on the application to protect.
In this work, we overcome this issue by proposing a machine-learning model that uses the CRS rules as input features.
Through training, ModSec-Learn is able to tune the contribution of each CRS rule to predictions, thus adapting the severity level to the web applications to protect.
Our experiments show that ModSec-Learn achieves a significantly better trade-off between detection and false positive rates.
Finally, we analyze how sparse regularization can reduce the number of rules that are relevant at inference time, by discarding more than 30\% of the CRS rules. We release our open-source code and the dataset at \url{https://github.com/pralab/modsec-learn} and \url{https://github.com/pralab/http-traffic-dataset}, respectively.

\end{abstract}
\setcounter{footnote}{0}
\section{Introduction}\label{sec:introduction}
Web applications are constantly evolving and deployed at a broad scale to offer a plethora or variegated services, imposing serious challenges in securing them against an increasing number of attacks \cite{web_sec_survey}.
Among these, \gls*{SQLi} consists of injecting a malicious SQL code payload inside regular queries, causing the target web application to either behave in an unintended way or expose sensitive data.
Even if countermeasures to this attack are well known~\cite{ml_driven,joshi2014sql,Kar2016Sqligot}, the \gls*{OWASP} Foundation still classifies \gls*{SQLi} as one of the top-10 most dangerous web threats~\cite{owasp_top_ten}.
Thus, \glspl*{WAF} are commonly used as a defense tool in enterprise systems~\cite{waf_survey,ml_driven} to counter such attacks and protect web applications.
They work by filtering the incoming requests directed towards the web applications and blocking suspicious connections.
In this work, we focus on \modsecurity~\cite{modsecurity_book}, an established open-source \gls*{WAF} solution that builds its defense on top of signatures of well-known attacks, collected by the OWASP Foundation and known as the \gls*{CRS}.
The \crs{} version used in this work (4.0.0) includes 319 rules, out of which 170 target critical injection attacks~\cite{crs_doc}.
Specifically, \gls*{SQLi} is the most represented class of injection attack counting 60 rules.
All rules are assigned with an heuristic \emph{severity level} used to evaluate whether an HTTP request is malicious or not.
Thus, detection is achieved through the summation of the scores of matched rules, blocking the incoming request if a threshold is exceeded.
However, this setup has three shortcomings: (i) the severity of each rule is purely heuristic, and it might not reflect the real behavior of the network to protect; (ii) rules only target attack patterns, but they are not taking into account legitimate network traffic, potentially yielding a high false positive rate; and (iii) different rules might either interfere with each other, or be redundant.

In this work we first show that the detection algorithm of \modsecurity based on \crs rules is largely ineffective due to the highlighted limitations.
We then propose \mlms, a novel machine-learning WAF that uses the CRS rules as input features.
In this way, the severity score of each rule is substituted by the weight attributed to that rule by the machine-learning model, allowing \mlms to tune their relevance and adapt itself to the web services to protect.
We test different machine-learning models used to implement \mlms, and we compare their predictive performance against ModSecurity, showing that the detection rate improves more than 45\% at 1\% false positive rate. 
Lastly, we investigate whether CRS contains redundant rules, computing embedded feature selection through $\ell_1$ regularization, highlighting that 18 out of 60 rules can be discarded as \mlms attributes no relevance on them.
\section{Background}\label{sec:background}
In this section we provide an overview of \gls*{SQLi} attacks and the \gls*{OWASP} CRS.

\myparagraph{SQL Injection (\gls*{SQLi}).}
It is a family of web threats that aims to retrieve sensitive information from a target database, modify data without authorization, or even execute privileged operations on the database~\cite{appelt2017tuningwaf}.
This can be achieved via specific SQL code fragments that are passed in the original request.
If the application does not sanitize the user-provided input and simply concatenates it with the rest of the query, the SQL fragment is interpreted as part of the original SQL command.
For example, considering the vulnerable SQL query of Listing~\ref{code:sql_query}, a malicious user could inject some SQL fragments in the \texttt{\$user} parameter, \emph{e.g.}, "\texttt{admin'--- }", to 
exfiltrate all the sensitive information of the provided username.
\newline
\begin{lstlisting}[
   language=SQL,
   basicstyle=\ttfamily,
   commentstyle=\color{gray},
   showspaces=false,
   showstringspaces=false,
   breakindent=1em,
   breaklines=true,
   % xleftmargin=1.8em,
   keepspaces=true,
   frame=single,
   caption={Example of SQL query vulnerable to SQL injection (SQLi).},
   captionpos=b,
   label={code:sql_query}
]
SELECT * FROM users WHERE username = '$user' AND password = '$passwd'
\end{lstlisting}


\myparagraph{ModSecurity.}
It is an open-source \gls*{WAF} solution for real-time web application security monitoring and hardening, which relies on a customizable set of rules to stop a large variety of web-based threats.

\myparagraph{The OWASP Core-Rule-Set (CRS) Project.}\label{sec:owasp_crs}
It is one of the most widely-used open-source sets of detection rules targeting OWASP Top 10 web security risks \cite{owasp_top_ten}.
It is widely adopted both in open-source \glspl*{WAF} like \modsecurity \cite{modsecurity_book} and Coraza\footnote{\url{https://coraza.io}}, as well as in more than ten commercial solutions~\cite{crs_doc}.

\mysubparagraph{Detection Rules.}
They are designed to detect specific types of web attacks.
Each rule is denoted by a unique identifier representing the specific class of attack it is intended to identify.
Rules are also associated with two notable configuration parameters, \ie, the \gls*{PL} and severity level.

\mysubparagraph{Paranoia Level.}
It is used to select which rules are enabled to analyze the HTTP requests \cite{crs_doc}.
The \crs includes four \glspl*{PL} (PL1 - PL4) and each rule is assigned to a specific \gls*{PL}.
Moreover, rules are grouped together by \gls*{PL} in a nested way: setting a certain \gls*{PL} enables all the rules assigned to that \gls*{PL}, as well as those assigned to lower \glspl*{PL}.
For instance, PL3 enables all the rules related to such \gls*{PL}, as well as those assigned to PL1 and PL2.

\mysubparagraph{Anomaly Scoring.}
Each detection rule is heuristically assigned with a \emph{severity level}, a positive integer value that quantifies how menacing a captured request is \cite{crs_doc}.
To compute a decision, \modsecurity applies the rules on incoming requests, and it sums all the severity levels of all the matches.
If such a summation exceeds a threshold, the incoming request is flagged as malicious.
In \crs there are four severity levels: CRITICAL (5), ERROR (4), WARNING (3) and NOTICE (2).
\section{Improving Modsecurity with Machine Learning}\label{sec:methodology}
\begin{figure*}[!t]
    \centering
    \includegraphics[width=0.9\textwidth]{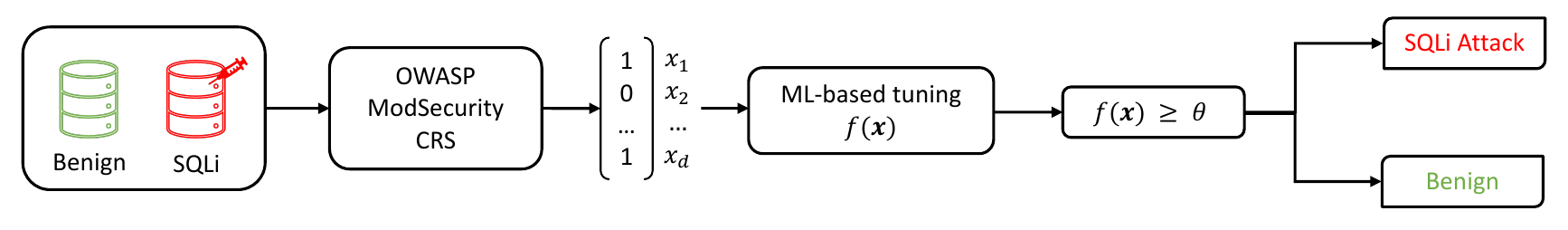}
    \caption{\mlms architecture. A machine-learning model is trained using the \crs rules as input features (52 features) to improve the trade-off between detection rate and false alarms. This amounts to learning a model of the incoming traffic directed towards the protected web services. Sparse regularization can also be used to select a subset of the available rules, instead of using PLs.}
\label{fig:system_overview}
\end{figure*}

We now present \mlms, the main contribution of our work, whose architecture is depicted in Fig.~\ref{fig:system_overview}.
\mlms is built on top of two main components: (i) a feature extraction phase that encodes the \crs rules into a vector representation, and (ii) a machine-learning model that learns how to optimally combine the \crs rules. 
This aims to surpass the shortcoming of manually tuning the severity levels while keeping the predictive power of the \crs rules.

\myparagraph{Detection Rules as Features.}
The input space is represented by SQL queries that are classified as malicious or benign by a machine-learning model.
Each SQL query is a string of readable characters, represented as $\vct z \in \set Z$, being $\set Z$ the space of all possible queries.
Let $\set D$ be the set of selected \gls*{SQLi} rules from \crs, and $\con d = |\set D|$ its cardinality.
We denote with $\phi : \set Z \mapsto \set X$ a function that maps a SQL query $\vct z$ to a $\con d$-dimensional feature vector $\vct x = ( x_{1}, \ldots, x_{\con d}) \in \set X = \{0,1\}^{\con d}$,
where each feature is set to 1 if the corresponding \gls*{SQLi} rule has been triggered by the SQL query $\vct z$, and 0 otherwise.
We want to remark that, although in this paper we focus on rules targeting SQLi attacks, this feature representation can be applied to any rule within the \crs.

\myparagraph{Optimal Combination of \crs Rules with ML.}\label{sec:ml_models}
We train three different machine-learning models on the aforementioned feature set. In particular, we use two linear models, i.e., \gls*{SVM} \cite{vapnik95} and a \gls*{LR} \cite{logistic}, using both $\ell_2$ and $\ell_1$ penalties; and a non-linear \gls*{RF} \cite{breiman2001rf} model.
Linear models are especially relevant in this context as they can be used to automatically tune the severity score to be assigned to each rule, instead of using the default ModSecurity values. Moreover, when using sparse ($\ell_1$) regularization, these models enable us to automatically select the optimal subset of CRS rules to be used, rather than resorting to a predefined PL.
Let us finally remark that our approach can be applied to any linear and non-linear machine-learning model, even if in the non-linear case it would be more complex to integrate the model within the existing ModSecurity implementation.

\myparagraph{Novel Dataset.} 
We aim to create a novel dataset consisting of legitimate samples based on real-world traffic as well as a comprehensive set of SQLi payloads.
Regarding the legitimate samples, since they are not readily available in the wild, we collected 508,529 samples provided in the open-appsec dataset\footnote{\url{https://github.com/openappsec/waf-comparison-project/tree/main/Data}}, which contains legitimate samples from various real-world scenarios.
However, the open-appsec dataset has only 458 SQLi payloads, which means that it is heavily biased towards legitimate samples.
To counter this issue, we augmented the original SQLi payload dataset of open-appsec using the following sources: (i) the HTTP Params dataset\footnote{\url{https://github.com/Morzeux/HttpParamsDataset}}, (ii) a SQLi dataset available on Kaggle\footnote{\url{https://www.kaggle.com/datasets/sajid576/sql-injection-dataset/data}} and (iii) a new set of SQLi payloads generated through the SQLi testing tool SQLmap\footnote{\url{https://sqlmap.org}} by executing SQLmap with different tampering scripts designed for payload obfuscation.
The final SQLi payload dataset includes 30,543 samples.
Finally, we created a balanced dataset by randomly selecting 25,000 benign and 25,000 SQLi samples from the novel dataset to ensure a fair evaluation of ModSecurity and machine-learning models.
\section{Experimental Analysis}\label{sec:experiments}
We now evaluate both \modsecurity (Sect. \ref{sec:exp-modsecurity}), showing that relying on heuristically assigned weights is suboptimal, and we continue by highlighting how \mlms enhances the performances thanks to the adaptation of weights (Sect. \ref{sec:exp-ml}), while also reducing the number of rules needed (Sect. \ref{sec:regularization}).

\subsection{Experimental Setup}\label{sec:exp_setup}
We now describe the setup underlying our experimental analysis.

\mysubparagraph{Training set} (\texttt{train}).
It contains 40,000 samples randomly chosen from the original dataset, divided in 20,000 benign and 20,000 \gls*{SQLi} queries to keep the two classes balanced.

\mysubparagraph{Test set} (\texttt{test}).
It contains 10,000 samples (5,000 benign, and 5,000 \gls*{SQLi} queries) randomly chosen from the original dataset.
This dataset has no intersection with the training set described above, and we use it to evaluate the performances of vanilla \modsecurity, and \mlms at different \glspl*{PL}.

\myparagraph{Setup of \modsecurity.}
We evaluate \modsecurity v3.0.10 with \crs v4.0.0, using \textit{pymodsecurity} v0.1.0\footnote{\url{https://github.com/AvalZ/pymodsecurity}}, which implements Python bindings to interface with \modsecurity.
Since we focus on the detection of \gls*{SQLi} attacks, we only enable the \gls*{SQLi} rules\footnote{\url{https://github.com/coreruleset/coreruleset/blob/v4.0.0/rules/REQUEST-942-APPLICATION-ATTACK-SQLI.conf}}. 
While the CRS consists of 60 rules, only 52 are activated by the samples in our training set; thus, we discard the remaining 8 rules.
These 52 are also the total number of features used to train models.

\myparagraph{\mlms with SVM, RF, and LR.}
We leverage the scikit-learn v1.4.0 \cite{sklearn} implementation of \gls*{SVM} (LinearSVC), \gls*{RF}, and \gls*{LR} to train each \mlms model. As for \mlms \gls*{SVM} and \gls*{LR}, we applied both $\ell_1$ and $\ell_2$ as penalization norms.
The \textit{saga}~\cite{defazio2014saga} solver was used to apply both norms to the \gls*{LR}.
We manually tested 5 values for the regularization parameter $C$ of SVM: $\{10^{-3}, 10^{-2}, 10^{-1}, 5 \cdot 10^{-1}, 1.0\}$. After training the \glspl*{SVM} and \glspl*{LR} for each \glspl*{PL} and penalization norms, we found that $5 \cdot 10^{-1}$ was the optimal value for the hyper-parameter $C$.
The other hyper-parameters were left to their default value.

\subsection{Evaluation of \modsecurity}
\label{sec:exp-modsecurity} 
\begin{figure*}[!t]
\centering
\includegraphics[width=1.0\columnwidth]{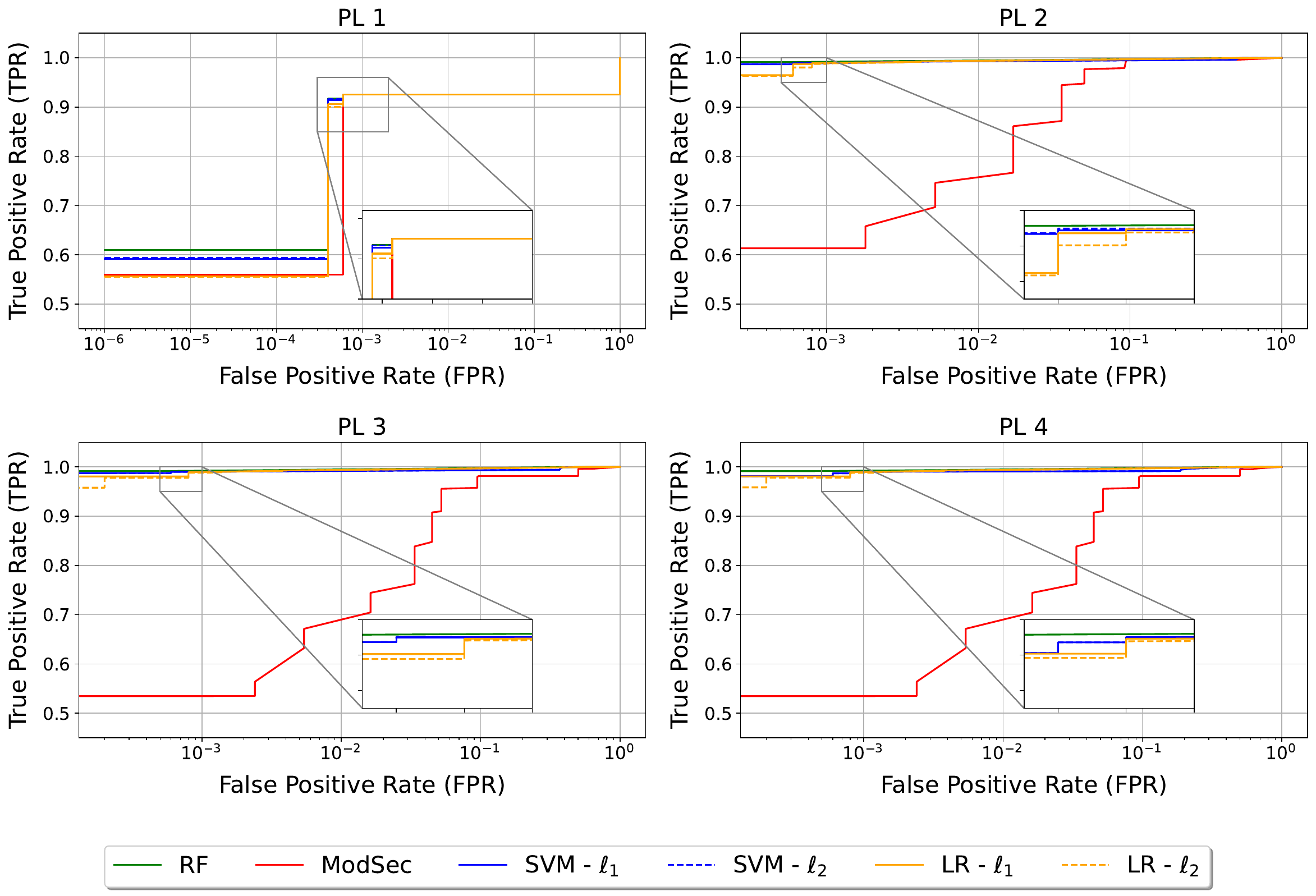}
\caption{ROC curves of \modsecurity vanilla (ModSec) and \mlms (SVM, RF, and LR), evaluated on \texttt{test}. Each curve reports the average detection rate of SQLi attacks (i.e., the True Positive Rate) against the fraction of misclassified benign SQL queries (i.e., the False Positive Rate). The zoomed section helps to understand the performance of each model when lines overlap.}
\label{fig:roc}
\end{figure*}

The first goal of our experimental analysis is understanding the detection capability of the vanilla \modsecurity. Rather than focusing only on its default values, we experiment with it over its entire configuration space, considering all the possible values for the \glspl*{PL} and the classification threshold. 
Hence, for each \gls*{PL}, we compute the Receiver-Operating-Characteristic (ROC) curve, which reports the True Positive Rate (TPR, i.e., the fraction of correctly-detected malicious SQLi requests) against the False Positive Rate (FPR, i.e., the fraction of wrongly-classified legitimate requests) obtained by considering all possible classification threshold values.
We report our findings with red lines in Fig.~\ref{fig:roc}, while in Table~\ref{tab:tpr_table}, we extrapolate the TPR values at 1\% FPR.
We would like to point out that, although the ROC curves in Fig.~\ref{fig:roc} already show the detection rates for each possible operating point (i.e., the value of FPR), we report the results in Table~\ref{tab:tpr_table} at 1\% FPR because it is a reasonable value commonly adopted in the literature \cite{corona2017deltaphish,Demontis2017YesML}.
We detail hereafter the key findings of our evaluations of \modsecurity against the test set (\texttt{test}).
The results of this first evaluation are indicated with red lines in Fig.~\ref{fig:roc}. 
The ROC curve of PL1 (default \gls*{PL} for \modsecurity) has proven to be the best among the \glspl*{PL} since, in this configuration, the number of active rules is minimal, only 20 rules enabled, causing a reduced number of false positives. 
The results for PL2 show a detection rate of 75.45\% at a 1\% FPR. This confirms that, as previously stated, having more active rules (31 more than PL1) leads to more false positives and decreased performance. Additionally, the ROC curves for both PL3 (which has 7 more rules enabled than PL2) and PL4 (which has 2 more rules enabled than PL3) are almost identical. This indicates that the additional \gls*{SQLi} rules of PL4 do not improve the detection capabilities.
Thus, as highlighted by the curves, at the best of its capabilities, \modsecurity with \crs is still missing plenty of potential threats.

\begin{table}[!t]
\caption{TPR at 1\% FPR of ModSec and \mlms (SVM, RF, and LR) evaluated on the test sets. For each WAF, we higlight the best results in bold.}
\label{tab:tpr_table}
\setlength{\tabcolsep}{5.5pt}
\renewcommand{\arraystretch}{0.5}
\centering
\begin{tabular*}{0.8\textwidth}{lllll}
\toprule
& PL1   & PL2   & PL3   & PL4   \\
\midrule
ModSec vanilla    &  \textbf{92.50}\%  &  75.45\%  &  68.55\%  &  68.55\% \\
\midrule
\mlms SVM ($\ell_1$)  &  92.50\%  &  \textbf{99.22}\%  &  99.04\%  &  99.02\% \\
\midrule
\mlms SVM ($\ell_2$)  &  92.50\%  &  \textbf{99.22}\%  &  99.04\%  &  99.02\% \\
\midrule
\mlms LR ($\ell_1$)  &  92.50\%  &  99.34\%  &  99.35\%  &  \textbf{99.35}\% \\
\midrule
\mlms LR ($\ell_2$)  &  92.50\%  &  99.34\%  &  99.34\%  &  \textbf{99.34}\% \\
\midrule
\mlms RF  &  92.50\%  &  99.41\%  &  99.45\%  &  \textbf{99.45}\% \\
\bottomrule
\end{tabular*}
\end{table}

\subsection{Evaluation of \mlms}
\label{sec:exp-ml}
We now analyze the performance of SVM, LR, and RF \mlms against the test set (\texttt{test}).
We plot the ROC curves in Fig.~\ref{fig:roc} using blue solid and dashed lines for \gls*{SVM} - $\ell_1$ and \gls*{SVM} - $\ell_2$, green solid lines for \gls*{RF}, yellow solid and dashed lines for \gls*{LR} - $\ell_1$ and \gls*{LR} - $\ell_2$, respectively.
They clearly show the superiority of \mlms w.r.t. the respective ModSecurity vanilla counterpart regardless of the operating point, i.e., for any FPR value, the detection rate of \mlms approaches is higher or equal for all \glspl*{PL}.
Specifically, considering the results of \gls{PL} 4 reported in Table \ref{tab:tpr_table}, the TPR at 1\% FPR of linear \gls*{SVM} with $\ell_1$ and $\ell_2$ is 44.71\% higher than \modsecurity. Considering the \gls*{LR} with $\ell_1$ and $\ell_2$, the TPRs at 1\% FPR is 44.93\% higher compared with \modsecurity.
While, as for the \gls*{RF}, the TPR at 1\% FPR is 45.07\% higher than the \modsecurity vanilla.
This confirms that, even by learning optimal weights, the rules enabled by PL1 are inappropriate for effectively discriminating benign samples from malicious ones.
Finally, unlike the \modsecurity vanilla, the majority of \mlms models achieve the best detection rate for PL4 (even though the results for PL4 are slightly higher than those obtained for PL2).
This result underlines that, even when adding rules that may lead to more false positives, machine learning can tune the importance of each rule to achieve a better TPR/FPR trade-off.

\begin{figure*}[!ht]
\centering
\includegraphics[width=0.9\columnwidth]{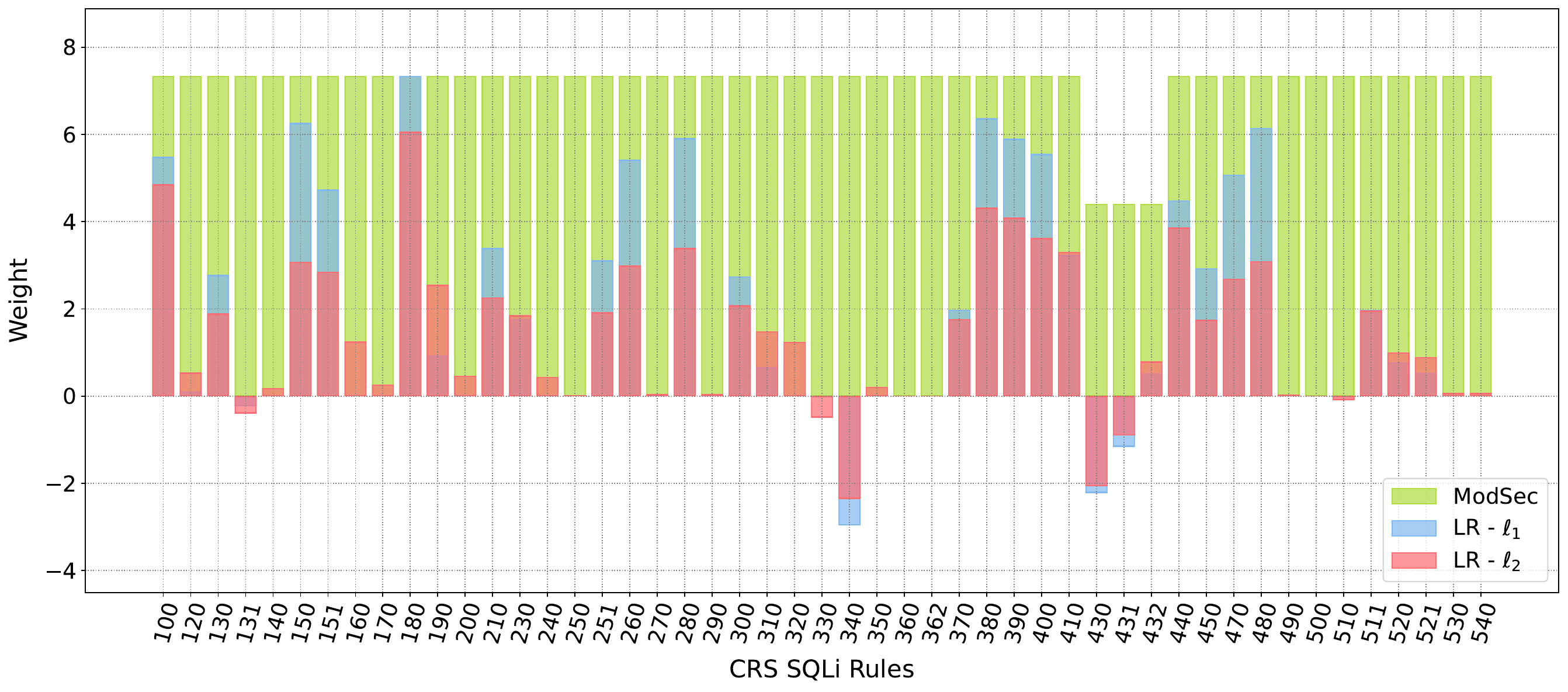}
\caption{Weight values learned at \gls*{PL} 4 by \mlms LR - $\ell_1$ (blue) and \mlms LR - $\ell_2$ (light red), and the weight used by \modsecurity vanilla (green). The additional color, i.e., red, is given by the overlapping of the green and blue bars with the light red ones. We only report the last three digits of the rule IDs on the x-axis as the first three digits are equal to 942 for all rules.}
\label{fig:lr_weights}
\end{figure*}

\subsection{Imposing Sparsity through Regularization}
\label{sec:regularization}
We now analyze the effect of regularization by investigating whether it is possible to select fewer rules from \crs as features. 
We leverage a regularization term with the $\ell_1$ norm to impose sparsity on the trained model, and we evaluate its impact on the relevance of each \crs rule on the classification process.
We also compare results with the weights computed through the inclusion of an $\ell_2$ regularization term.
Fig.~\ref{fig:lr_weights} displays the distribution of rule weights of \mlms implemented with LR at \gls*{PL} 4.
We chose this \gls*{PL} to enable all rules and provide a complete overview of their impact.
The blue and light red bins represent the weights computed with $\ell_1$ and $\ell_2$ regularization, respectively, while the green ones are the \modsecurity severity scores (overlaps are colored in dark red). Since severity scores ranges from 2 to 5, we normalize them using the minimum and maximum of \gls*{LR} ones.
The results presented in Table~\ref{tab:tpr_table} demonstrate that the $\ell_1$ regularization norm can achieve the same performances of \gls*{LR} with $\ell_2$ norm while using even fewer rules, 18 rules weighted as 0, while $\ell_2$ norm and \modsecurity use all available rules. It's important to note that the rules set to a weight of 0 by the machine learning model are deemed unnecessary for the classification task. Moreover, some rules might receive negative weights, indicating that their presence is more indicative of legitimate behavior.
Applying this approach to the configuration of security tools such as \modsecurity introduces a more grounded and less arbitrary method for security rule selection. 
Rather than relying on manual selection or a predefined set of rules that may not be optimal with respect to the data that will then be found to classify, the use of \mlms makes it possible to automate both the selection of rules and the assignment of their weights, optimizing \modsecurity performance on the data.

\section{Related Work}\label{sec:related_work}
Although previous work has proposed several machine-learning solutions to counter SQLi attacks \cite{joshi2014sql,Kar2016Sqligot}, this study focuses specifically on ModSecurity. Earlier research \cite{singh2018_ms_pl_study,sobola_modsecurity_study} evaluated \modsecurity's performance, considering the impact of the \gls*{PL} under various web attacks, but used limited attack samples and did not analyze the TPR-FPR trade-off. Folini \etal~\cite{crs_ml} explored unsupervised anomaly detection for unknown attacks, while Tran \etal~\cite{Tran2020ModSecML} propose a first attempt to combine machine learning with the \crs{} rules. However, they did not evaluate \modsecurity.
Nguyen \etal,~\cite{nguyen2022improvingwaf} proposed a hybrid approach, combining \modsecurity with a machine-learning model for request categorization. However, none of these studies assessed the TPR-FPR trade-off for each \gls*{PL} as in this work.
Furthermore, we are the first to investigate the effectiveness of regularization for selecting useful rules. Finally, we also share our dataset to foster future research.

\section{Conclusion and Future Work}\label{sec:conclusions}
In this work we propose \mlms, a novel methodology for training machine-learning classifiers using the \crs rules as input features.
This permits the adaptation of the severity levels of \crs rules to the application to defend, achieving the best trade-off between detection and false positive rates.
We show that \mlms improves the detection rate of the vanilla \modsecurity by more than 45\%, while removing more than 30\% of the CRS rules through embedded feature selection with $\ell_1$ regularization.
While we only target \gls*{SQLi} attacks in this work, we argue that our methodology is general enough to tackle also other web threats, and it can be applied "as is" on different rulesets.
Furthermore, even if we only focused on \modsecurity, our evaluation can be repeated also on other open-source and commercial WAFs. 
To conclude, we firmly believe that our work will pave the way towards strengthening classical rule-based solutions with machine learning-based approaches, filling the gap between these two worlds.

















\begin{credits}
\subsubsection{\ackname} This research was supported by the TESTABLE project, funded by the European Union's Horizon 2020 research and innovation program (grant no. 101019206); and the ELSA project, funded by the European Union's Horizon Europe research and innovation program (grant agreement no. 101070617); and the SERICS project (PE00000014) under the MUR National Recovery and Resilience Plan funded by the European Union – NextGenerationEU.
\end{credits}
%
%
%
\bibliographystyle{splncs04}
\bibliography{bibliography}

\end{document}